# Deep Reinforcement Learning Xiangqi Player with Monte Carlo Tree Search

**Junyu Hu**  jh4930@columbia.edu

Department of Electrical Engineering, The Fu Foundation School of Engineering and Applied Science, Columbia University, New York, NY 10027, USA

**Jinsong Liu**  jl6850@columbia.edu

Department of Electrical Engineering, The Fu Foundation School of Engineering and Applied Science, Columbia University, New York, NY 10027, USA

**Berk Yilmaz**  by2385@columbia.edu

Department of Electrical Engineering, The Fu Foundation School of Engineering and Applied Science, Columbia University, New York, NY 10027, USA

## Abstract

This paper presents a Deep Reinforcement Learning (DRL) system for Xiangqi (Chinese Chess) that integrates neural networks with Monte Carlo Tree Search (MCTS) to enable strategic self-play and self-improvement. Addressing the underexplored complexity of Xiangqi—including its unique board layout, piece movement constraints, and victory conditions—our approach combines policy-value networks with MCTS to simulate move consequences and refine decision-making. By overcoming challenges such as Xiangqi's high branching factor and asymmetrical piece dynamics, our work advances AI capabilities in culturally significant strategy games while providing insights for adapting DRL-MCTS frameworks to domain-specific rule systems. Our project github link can be found at: https://github.com/JL6850/E6892-FinalProject

## 1. Introduction

Our research introduces a Deep Reinforcement Learning system that learns to play Xiangqi(Chinese Chess) through strategic self-play and self-improvement. The approach combines two components: neural networks that guide decision-making and evaluate positions, paired with Monte Carlo Tree Search(MCTS) that simulates different moves and their consequences. Through repeated gameplay against itself, the system gradually develops effective strategies while accounting for Xiangqi's distinctive rules and gameplay dynamics. Our methodology draws inspiration from modern game-playing systems like AlphaGo, but specifically addresses the technical challenges posed by Xiangqi's unique pieces, board layout, and victory conditions.

### 1.1 Background information

Xiangqi is a longstanding traditional Chinese strategy board game that has existed for centuries and has millions of players, primarily in East and Southeast Asia. Similar to Western Chess, Xiangqi is a two-player game of perfect information, but there are many things that are different in Xiangqi which contribute to Xiangqi being a more complex variant. Xiangqi is played on a 9×10 board with 90 positions. the 8×8 chess board with 64 squares, resulting in a larger and more irregular state space. In addition, pieces in Xiangqi move in asymmetrical ways: Advisors and Elephants, for example, can only occupy specific areas such as the palace or a designated side of the river. The Cannon introduces its own capturing mechanism, and we have the added stipulation that the two Generals cannot face each other which imposes tactical constraints. Factors such as these lead to a greater average branching factor and longer average game length in Xiangqi. An overview of complexity indicators comparing Xiangqi and Chess can be found in the table below:

| METRIC | XIANGQI | CHESS |
| --- | --- | --- |
| BOARD SIZE | $9 \times 10$ | $8 \times 8$ |
| STATE SPACE | $\sim 10^{40}$ | $\sim 10^{47}$ |
| GAME TREE | $\sim 10^{150}$ | $\sim 10^{123}$ |
| BRANCHING FACTOR | $\sim 38$ | $\sim 35$ |
| GAME LENGTH | $\sim 95$ | $\sim 70$ |

Table 1: overview of complexity indicators

## 2. Related Work

### 2.1 Overview

The combination of DRL and MCTS, first successfully implemented in AlphaGo and AlphaZero, has completely changed the way AI makes decisions in chess games. This type of method combines the judgment of the situation by the neural network with the search algorithm, and continuously improves the accuracy of move prediction and the ability to judge the pros and cons of the situation through self-play. Although originally designed for Go and chess, its success has also led to many applications expanding to real-time strategy games such as StarCraft II, as well as broader complex systems such as robot control and resource scheduling.

In contrast, Xiangqi AI relies more on rule-driven methods, such as alpha-beta pruning and human-designed evaluation functions. These systems perform well in efficiency, but lack adaptability when faced with complex or novel situations. In recent years, some work has introduced neural networks to predict moves or evaluate situations, but most of them only focus on a certain link and lack the overall planning and exploration capabilities brought by MCTS. Although the DRL model performs well in learning local move patterns, due to the many branches and long tactical chains of chess, it is difficult to achieve effective long-term planning by DRL alone; and pure MCTS is easy to get stuck in the huge search space, and the computational overhead is extremely high. Therefore, in order to achieve powerful AI in chess, it is necessary to combine the advantages of DRL and MCTS more closely.

## 3. Methods

### 3.1 Alpha–Zero Model For International Chess

Before implementing the Xiangqi version implementation we wanted to try the classical Alpha-Zero chess model to get familiar with the environment and data loading setup. The biggest difference between this classical chess implementation and our Xiangqi player is that in the Xiangqi environment we utilize a training set to train our neural network model to get a more advanced version of our player. In order to get a comprehensive version we first defined the international chess environment, then evaluated a neural network policy with the values. We modified the move encoder and utilized the Monte Carlo Tree Search algorithm for the decision making process. After that we evaluated our model by self-playing by itself and with the data it generated we improved our model.

#### 3.1.1 Setting up the Game Environment

To set up the game environment we first defined a ChessState where we utilized the python-chess library. Python-chess library enabled us to get legal moves, simulate the moves, and determine the chess game status whether it is a win situation. In this environment, we convert the board sensor to tensor for our neural network architecture. Unlike the Xiangqi environment, in international ches we define 13 states, which 12 comes from the pieces and 1 is extra to determine for whose turn it is.

#### 3.1.2 Representation of Network Channels

We utilized the following channel structure for our international chess implementation

| Channel | Representation |
| --- | --- |
| 0-5 | White pieces |
| 6-11 | Black pieces |
| 12 | Turn indicator |
| 13 | White kingside castling rights |
| 14 | White queenside has the castling |
| 15 | Black kingside castling rights |
| 16 | Black queenside castling rights |
| 17 | En passant square |
| 18 | Check indicator |

Table 2: Chess implementation channel structure

#### 3.1.3 Neural Network: Policy + Value Network

Then we defined the neural network and policy for our international chess implementation. For our neural network architecture we followed the Alpha-Zero style structure. Alpha-Zero implementation consists of "dual-head" convolutional network implementation where the network is divided into three pieces. The shared body has an input shape of [batch_size, 19, 8, 8] where each of the 19 "planes" encodes one bit‑plane of the board. Then we increase the hidden dimension size from 19 to 256 by introducing a simple 3x3 convolutional layer followed up by batch normalization and ReLU to normalize and standardize the weights. The output from the convolutional layer gives us the raw embeddings into high dimensional feature space (256). After the 3x3 convolutional block we introduce residual block architecture. Although the original paper states that they are stacking 19–39 residual blocks because of the limited computational power we finalized with 10 residual blocks to lighten the environment. Each of our residual blocks consist of a convolutional layer followed by batch normalization and ReLU, after the first convolutional block we introduce another convolutional layer and batch normalization. After these steps, we introduce the original input one more time as a residual connection and we perform ReLU on the sum. The skipped connections helped us prevent the vanishing gradients problem while we are training.

#### 3.1.4 Policy Head

Our policy head starts working with the information it received from the shared body which includes 256 channel-feature maps. After receiving the weight information we utilize another 3x3 convolutional layer without decreasing or increasing the hidden layer dimensions. After the layer, we flatten the 256×8×8 feature maps into one long vector because the final step in the policy head is a fully-connected (linear) layer, and those layers expect a 1D input. Although the original paper introduces 16384 dimensional vectors, due to GPU and memory constraints we have hardcoded a dimension size of 42. Later, we used these dimensions and applied Softmax to get the next move probabilities.

### 3.1.4 Value Head

Our value head consists of one 1x1 convolutional layer which has an output of 1, by implementing this strategy we are reducing the hidden channels of our structure to 1 then we follow up with batch normalization and and ReLU and afterwards flatten the [8,8] dimensions to get a 64 dimensional vector. After converting the dimension size to 64, we expand it with a linear layer to its original form of 256 and then we apply a final layer to reduce the hidden dimensions to 1 again. Our value head consists of tanh function so we can get the probability values in range [-1,1], representing the loss with -1 and win with 1.

### 3.1.5 Move Encoder

In order to continue with the Alpha-Zero implementation we need to introduce bi-direcitonal mapping to our environment. Bi-directional mapping is required in our project because the network doesn't output a list of moves. It gives out a fixed-length vector of raw scores or probabilities whose positions correspond to "move indices." The move encoder tells us which vector slot means the corresponding piece. Move encoder helps us map each legal move to a unique index number so that we can check the rules immediately. In order to achieve this, we create a one-hot-encoding output vector where we can get the probabilities. Our move encoder also checks and handles the promotions, castling and some advanced rules like en-passant. In move encoder, we utilize three main functions, encode_move(move) returns an index value for each move as we stated before, decode_move(index) does the exact opposite thing and returns us a state from the given index, we also utilize policy_to_moves() function to filter the neural network outputs for legal moves.

### 3.1.6 MCTS (Monte Carlo Tree Search)

As also used in the main implementation of Alpha-Zero model, Monte Carlo Search Tree algorithm is used to generate a search strategy. The moves are selected by not only getting the highest-probability actions every time, we further implement this algorithm by adding selection and giving room for exploration. The policy refers to which moves are going to be explored and value on the other hand refers to the estimation of how good a leaf state is in the current state. We first begin our implementation by maximizing the Upper Confidence Bound (UCB)

$$UCB(s,a) = Q(s,a) + c * P(s,a) * \frac{\sqrt{N(s)}}{1+N(s.a)}$$

where, Q(s,a) = **average value** of taking action a at state s, P(s,a) is the **prior probability** from the policy network, N(s) is the **total visit count** to node s, N(s,a) is how many times child a was visited and c is the exploration constant which we change during training to evaluate the scores. At the end of each iteration we pick the child note with the highest upper bound confidence level. And once we complete the iteration and receive a leaf we then update the policy and value from the neural network and for each legal move we add a child node to that tree with prior initializations N(s,a)=0 , Q(s,a)=0 and P(s,a)= p which we got from our neural network. The difference between traditional Monte Carlo Tree Search and Alpha-Zero implementation is that in standard MCTS we continue to simulate the entire game but Alpha-Zero model directly gets the value(v)* from the neural network. (*v = ∈[−1,1] ). After we receive our value we start the backpropagation. For each node in each iteration, we first update the visit count: N(s,a) ← N(s,a) + 1, total value: W(s,a) ← W(s,a) + v and mean value Q(s,a)←W(s,a) / N(s,a) and each time while we do the updates we flip the sign of our value (v) because of the turn rule. After we initialize the structure we can simulate it and estimate a policy with the equation below:

$$\pi(a) = \frac{\sqrt{N(s,a)}}{\sum_b N(s,b)}$$

which we get the improved policy.

### 3.1.7 Self-Play

Like the Alpha-Zero paper in order to train our data we utilized a self-playing algorithm to collect and then update our model. In this section, the model has played games against itself and generated data. For each move we first used Monte Carlo Search Tree to generate a policy distribution then we stored that policy distribution in dimensions of (state, policy, value_placeholder), then by looking at the stored policy we made a move. If our is_end_game function starts we assign a value to each position in the table based on the final outcome of the game.

### 3.1.8 Conclusion of Alpha-Zero for Chess

The Alpha-Zero for international chess implementation gives us some important decision making ideas to improve the system and convert it to the Xiangqi environment. We managed to train the Alpha-Zero model using A100's in 4 days. As the model's complexity increased the games started to last longer, In first iterations games were completed in 30 minutes whereas in Iteration 10 each game lasted for an hour. Due to limited constraints we stopped training at iteration 10, but our model shows clear improvement and can be further improved in future iterations.

### 3.2 Model for Xiangqi

As we progressed in Alpha-Zero algorithm, in this section we will discuss the changes in the environment and improvement we have made during the training. Alpha-Zero algorithm does not utilize any pre-training steps and collects the data by self-play. However, as we have seen in the results the early models lack any technical abilities and strategies. In order to prevent this, we collected a dataset with a script containing previous Xiangqi games, which will be discussed in more detail in section 3.2.2.

#### 3.2.1 Introduction to Xiangqi Rules

First, we introduce the basic piece-movement rules of Xiangqi to establish a clear understanding of what constitutes a legal move (see Table 3). Next, Xiangqi incorporates a number of special movement constraints such as the elephant's inability to cross the river, the cannon's requirement to leap over exactly one intervening piece when capturing, palace-bound advisors and generals, and the prohibition against generals facing each other directly—which serve to evoke aspects of ancient battlefield tactics and logistics. Finally, winning in Xiangqi, like in international chess, is achieved by capturing the opposing general . However, at high levels of play a substantial proportion of games end in draws, owing largely to Xiangqi's unique draw mechanisms such as perpetual check limitations, drawn-position rules when neither side can make progress, and repetition-based adjudication—which collectively reduce the likelihood of decisive outcomes.

| Name | Movement |
| --- | --- |
| General | Moves one square |
| Advisor | Moves one square diagonally |
| Elephant | Moves two squares diagonally |
| Horse | Moves two square orthogonally then one square diagonally (leg block applies) |
| Rook | Moves any number of squares orthogonally |
| Cannon | Moves any number of squares orthogonally |
| Soldier | Moves one square forward |

Table 3: Xiangqi's legal move

#### 3.2.2 Xiangqi Dataset Setup

There are a limited number of libraries that contain the Xiangqi environment. However, throughout our project we utilized OpenAI's Xiangqi environment. We created a script that creates a game environment using the OpenAI Gym framework and gym_xiangqi package. By using this environment we could access the observation space, action space, the step functions which we execute to make moves and return new state, action, reward and game status information and reset functions which we used to start a new game. Our dataset adopts the Portable Game Notation (PGN) format. The [FEN] tag is used to encode the initial board configuration, while [Result] and [Format "ICCS"] are retained to capture the game outcome and move history, there were a total of 141,484 games recorded with 11,683,480 total moves. Among these games red won 53,787 times whereas blacks won 39,884 times and 47,843 of the games were concluded as draws.

#### 3.2.3 Supervised Pre-training from Historical Games

In the original AlphaZero framework, training starts completely from scratch: both the policy and value networks are randomly initialized and immediately paired with MCTS for self-play. At this early stage, the untrained networks produce near‑uniform move priors, and MCTS adds heavy Dirichlet noise to promote exploration, so the first several games effectively resemble random play. Converging from such a cold start typically requires tens or even hundreds of thousands of high‑quality self‑play games, demanding computational resources and training time far beyond our means.

To mitigate this, we introduce a supervised pre training phase before the self‑play module. By leveraging existing human game records and using behavior cloning, we train an initial policy–value network that imitates expert moves. This yields a substantially more informed starting strategy than random initialization.

##### 3.2.3.1 Preparation Before Training

During the feature‑encoding phase, we map the Xiangqi board and side to move into a multi‑channel binary tensor of shape (10×9×15): for each non‑empty square, we set the corresponding piece's channel to 1 (and all other piece channels to 0), thereby assigning each of the seven Red piece types and seven Black piece types to the first 14 channels; the fifteenth channel is uniformly set to indicate which side is to move (Red or Black) across the entire board. This representation both preserves the spatial distribution of pieces and explicitly encodes move‑ownership.

For action encoding, since the Xiangqi board has 10×9=90 squares and each move is essentially "from one square to another," there are 90×90=8100 possible source–destination combinations. We compute a unique action index in [0,8099] by taking (source_index×90 + destination_index). As a result, the policy network needs only output a probability vector of length 8100 to cover every possible move, creating a one‑to‑one correspondence between network outputs and legal actions.

### 3.2.3.2 Model Construction and Training

In the model construction phase, we based our design on the AlphaZero–style dual‑head convolutional neural network, dividing the workflow from input to output into three major components:

(i) Input and Feature Extraction Backbone
The network accepts a tensor of shape (10, 9, 15) as input. First, a 3×3 convolutional layer with 256 filters is applied, followed by batch normalization and a ReLU activation, which maps the sparse binary inputs into a high-dimensional continuous feature space to capture initial spatial correlations. We then stack several residual blocks, each composed of two 3×3 convolutions + batch normalization + ReLU. A skip connection adds the block's input back into its output before a final ReLU.

(ii) Policy Head
From the shared backbone, one branch first applies a 1×1 convolution to reduce channel dimensionality, then flattens the resulting feature maps into a 1D vector. A final fully connected layer projects this vector into 8,100 output nodes, and a Softmax activation converts these into a probability distribution over all legal moves. Each output index corresponds to one "source‑square × destination‑square" combination, so the network's output directly represents the likelihood of selecting each possible move.

(iii) Value Head
The other branch similarly begins with a 1×1 convolution to condense channels, then flattens the feature maps into a 64-dimensional vector. This vector passes through a fully connected layer that expands it to 256 dimensions, and finally through a single-neuron output layer with a tanh activation. The resulting scalar lies in [−1, 1], representing the estimated game outcome from the current side-to-move perspective (−1 = certain loss, +1 = certain win).

During training, we jointly optimize the policy and value heads by combining their loss functions: sparse categorical cross-entropy for the policy head and mean squared error for the value head, weighted equally (1:1). We use the Adam optimizer for its adaptive learning-rate properties, which enable rapid initial convergence and stable training. We also monitor policy accuracy and value MAE on a held-out validation set to assess move‑prediction quality and value‑estimate precision. The result is a model with a solid foundational move policy (e.g., it knows how to capture and check), which we then employ in the subsequent self‑play stage.

### 3.2.4 Self-playing in Xiangqi

After completing the supervised pretraining for Xiangqi, we employed the same self-play methodology used in international chess: integrating the pretrained policy–value network with MCTS to generate self-play games, iteratively collecting new state–policy–value triplets, and retraining the network to further improve its performance.

## 4. Results

### 4.1 Evaluation of International Chess Alpha-Zero Model

In this part we analyze games and go through each training and changes in our model.

```
Initialized move encoding with 42 possible moves
Starting iteration 1/10
Generating self-play games...
Starting self-play game 1/10
  Move 10, State: r1bqkbnr/p1p1ppp1/7p/1pnp4/P4P1P/3P2P1/1PP1P3/RNBQKBNR w KQkq - 1 6
  Move 20, State: 2bqkbnr/p1p1p3/1r3ppp/1p1p4/P4P1P/1nPPP1PN/1P2B3/RNBQK1R1 w Qk - 2 11
  Move 30, State: 4kbnr/pb2p3/r4ppp/qpp5/P2p1P1P/NnPPP1PN/1P2B3/1RBQ1KR1 w k - 2 16
  Move 40, State: 4kbn1/p3p2r/r1b2p1p/q1p3Qp/P2p1P1P/pnPPP1PN/1P2K3/1RB3R1 w - - 2 21
  Move 50, State: 4kbnr/p3p3/5p1Q/q6p/Pr1P1P1P/pn1PK1PN/1P4b1/1RBR4 w - - 3 26
  Move 60, State: 1r1q1bnr/p2kp3/4bp2/7p/P2PQP1P/pn1PK1PN/1P6/1RBR4 w - - 13 31
  Move 70, State: 3q1bnr/p2k4/5p2/nPr1p2p/P2P1PbP/p2PK1PN/4Q3/1RBR4 w - - 0 36
  Move 80, State: 3q2nr/p2k4/3b4/BPr2p1p/P2pKPbP/p2P2PN/6Q1/1R3R2 w - - 0 41
  Move 90, State: 6nr/p1qk4/3b4/BP3p1p/P1rK1PbP/p2P1QPN/8/1R3R2 w - - 9 46
  Move 100, State: 7r/pq1k4/3b1n2/BP3p1p/P1r2PQP/p2PK1Pb/8/1R3R2 w - - 8 51
  Move 110, State: 4k2r/pqr1b3/5n2/BP3Q1p/PR3P1P/3PK1Pb/p7/5R2 w - - 0 56
  Move 120, State: 4k2r/p1r1b2n/1Pq5/B6p/P3QP1P/3PK1Pb/p7/2R2R2 w - - 9 61
  Move 130, State: 4k2r/p3b2n/1P2q3/B2Q3p/P4P1P/3P1KPb/8/n2rRR2 w - - 8 66
  Move 140, State: 4kr2/p3b3/1P3n2/B6p/P3RPqP/3P2Pb/Q4K2/n2r1R2 w - - 18 71
  Move 150, State: 3k1r1q/p2nb3/1P6/7p/P3RP1P/Q2P1KPb/8/n2rBR2 w - - 28 76
Game finished with outcome: 0
```

Figure 1. Example of self-play data

The figure above shows the moves in each state, as stated before the policy network has not been trained and it is first randomly initialized. MCTS priors are based on noisy policy outputs, in this period there is no value network guidance yet. Also, there is no information about the past games since this is the first training iteration. The game lasted for 76 moves which means our model is exploring the options on the board, the game is very low-level, there are no tactics involved , no evident pursuit of checkmate or material gain, and moves seem arbitrary. At move 150, the final board is 3k1r1q/p2nb3/1P6/7p/P3RP1P/Q2P1KPb/8/n2rBR2, double queens, imbalanced pawns, centralized minor pieces with no plan which shows the random play.

However as we progress in the training as shown in the figure below

```
Training neural network...
Epoch 1/5, Loss: 0.8124, Policy Loss: 0.8086, Value Loss: 0.0039
Epoch 2/5, Loss: 0.8317, Policy Loss: 0.8310, Value Loss: 0.0006
Epoch 3/5, Loss: 0.6145, Policy Loss: 0.6141, Value Loss: 0.0004
Epoch 4/5, Loss: 0.5511, Policy Loss: 0.5511, Value Loss: 0.0001
Epoch 5/5, Loss: 0.5356, Policy Loss: 0.5356, Value Loss: 0.0000
Completed iteration 1
Starting iteration 2/10
Generating self-play games...
Starting self-play game 1/10
  Move 10, State: rnb1kb1r/pp3ppp/1qp1pn2/3p4/5P2/1PP1P3/P2PB1PP/RNBQK1NR w KQkq - 1 6
  Move 20, State: rnb1kb1r/pp1n1pp1/1qp1p3/3p3p/3P1P2/NPP1P2P/P3B1P1/1RBQ1KNR w q - 5 11
  Move 30, State: rnb1kb1r/pp1n1pp1/2p5/1Bqpp2p/3P1P2/PPP1PNPP/8/1RBQ1KR1 w kq - 0 16
  Move 40, State: r3kbnr/p2n1pq1/1p5p/4p1p1/1p2P1P1/P1PP1PPN/4B3/RN3KR1 w kq - 1 21
  Move 50, State: r1k2r/p2n1p2/1pp5/1B1pp1pp/1b1P1P2/PP1QPNPP/4K3/1RB2R2 w q - 0 21
  Move 60, State: r1b1k2r/p2n3Q/1p2p3/1p5p/1b1p1P2/PP2PpPP/1B2K3/1R4R1 w q - 0 26
  Move 60, State: r1b1k2r/p2n3Q/np2p3/1p3P2/8/bP2PppP/1B4R1/1R3K2 w q - 0 31
  Game finished with outcome: -1
```

Figure 2. Training feedback of the first iteration

We got a loss of 0.5336, which is the combination of policy loss (0.5336) and value loss (0.0). After the first iteration our game shows some improvements. Based on looking at the FEN boards, even if it's after the first iteration we can see the correct pawn structures from both sides. We can see our model started to use other pieces such as bishops to take control of the board which can be seen in movements 3P1P2, PPP1PNPP which indicates there is better piece coordination. However, the model still does blunders such as in move 30 the player sets its black queen to g2 which is a clear blunder and technical error. In this game black shows more aggression and at the end wins the game.

```
Collected 1364 training examples so far
Training neural network...
Epoch 1/5, Loss: 0.5736, Policy Loss: 0.5199, Value Loss: 0.0537
Epoch 2/5, Loss: 0.5561, Policy Loss: 0.5042, Value Loss: 0.0518
Epoch 3/5, Loss: 0.6032, Policy Loss: 0.5502, Value Loss: 0.0530
Epoch 4/5, Loss: 0.6011, Policy Loss: 0.5495, Value Loss: 0.0516
Epoch 5/5, Loss: 0.5495, Policy Loss: 0.5012, Value Loss: 0.0483
Completed iteration 2
Starting iteration 3/10
Generating self-play games...
Starting self-play game 1/10
  Move 10, State: rn1qkbnr/2p1pppp/1p6/p2p4/6b1/P1P1P1P1/1P1PQP1P/RNB1KBNR w KQkq - 0 6
  Move 20, State: rn2kbnr/2p2p1p/1pq5/p2pp1p1/1PQ3P1/P1P1P1P1/3PBP2/RNB1K1NR w KQkq - 1 11
  Move 30, State: r3kbnr/2p2p2/1pn4p/p3pqp1/1Pp1P1P1/P1PP1PPN/4B3/RNB2K1R w kq - 0 16
  Move 40, State: r3kbnr/2p1npq1/1B5p/4p1p1/1p2P1P1/P1PPN/4B3/RN3KR1 w kq - 1 21
  Move 50, State: 2r2bnr/2p1kpq1/7p/3np1p1/3BP1P1/P1Pp1PP1/1p2BNK1/RN4R1 w - - 2 26
  Move 60, State: 2r2bnr/B3kp1q/2p4p/1q1Pp1p1/5PP1/P1Pp2P1/4B3/3N2RK w - - 0 31
  Move 70, State: 6nr/4kpbq/2q4p/r3p3/5pP1/P1PB2P1/5B2/3N2RK w - - 0 36
  Move 80, State: 6nr/4kpbq/r4q1p/4pB2/5pP1/P1P1B1PK/8/3N2R1 w - - 10 41
  Game finished with outcome: -1
Starting self-play game 2/10
  Move 10, State: 1nbqkbnr/rp1p1ppp/8/p3p3/2p5/P1P1P1P1/1P1PBP1P/RNBQK1NR w KQk - 0 6
  Move 20, State: 1n1qkb1r/rp2nppp/3p4/p3p3/P7/1pPbPPPP/4B3/RNBQK1NR w KQk - 0 11
  Move 30, State: 1n1q1b1r/rp1k1ppp/3p4/pn2p3/P3P1P1/NQP2P1P/2b1B3/1RB1K1NR w K - 1 16
  Move 40, State: 1n5r/rpnk1ppp/3p1b2/p3p3/P3P1P1/NQP1BP1R/2b1B3/1R3KN1 w - - 3 21
  Move 50, State: 1n1kr3/rp3pp1/3pnb1p/p3p3/P3P1P1/NQP1BPR1/4B3/5KN1 w - - 2 26
  Move 60, State: 1n2rn2/r3kpp1/1p5p/2pp1b1/P3PBP1/NQP1KPRN/8/3B4 w - - 2 31
  Move 70, State: 1n1r4/r4ppn/1p2k2p/2pp3/P1B1P1P1/NQP2PRN/5K2/2b5 w - - 8 36
  Move 80, State: 8/r2n1prn/1p3k2p/2pp2pp3/PNB1PPP1/2P3RN/6K1/1Qb5 w - - 1 41
  Move 90, State: 5n2/r2r3n/1p2k1pp/p4p2/P1p1PpPN/2P3R1/2N5/1Qb3K1 w - - 0 46
  Move 100, State: 5n2/rr5n/1p1k2pp/p4p2/P1pN1pP1/2P1N1R1/8/1Q4K1 w - - 4 51
  Move 110, State: 5n2/rr1k4/6pp/p2n1p2/P1p1Qp1/2P1N1R1/4N3/5K2 w - - 3 56
  Move 120, State: r7/1r5n/3k1n1p/p4p2/P4pP1/2P5/4N1R1/5K2 w - - 2 61
  Move 130, State: r7/1r5n/4K2P/p4p2/P4p2/2Pn4/4N1R1/5K2 w - - 5 66
  Move 140, State: 2r5/r6n/3k3P/p4p2/P7/2PnK2N/3R4/8 w - - 9 71
  Move 150, State: 2r5/r1k5/7P/p4p2/P3n3/2P1K2N/3R4/8 w - - 9 76
  Game finished with outcome: 0
```

Figure 3. Training feedback of the second iteration

The figure above shows 2 games after iteration 2. After iteration 2 our loss drops to 0.5495 and value loss starts to increase which indicates that the value head is trying to evaluate the board states. Our policy loss is stable and consistent with iteration 0 and iteration 1. First game lasts for 80 moves which at the end black wins again. In early stages of the game the model still makes some randomized movements and starts to explore the board, However, around the middle game both players show central play coordinations and tactics such as putting their rooks and queens to the open files. This game shows the model is trying to evaluate a strategy based on the prior movements. In the second game, we observe something different, in late early game players castle and for the first time the model starts to enter the late-game stages which can be seen looking at this FEN line "3n3/2r5/k3KP/p4P2/P3n2/8/8/7R" which shows piece counts and pawn play.

```
Collected 1500 training examples so far
Training neural network...
Epoch 1/5, Loss: 0.0725, Policy Loss: 0.0661, Value Loss: 0.0064
Epoch 2/5, Loss: 0.0644, Policy Loss: 0.0589, Value Loss: 0.0055
Epoch 3/5, Loss: 0.0642, Policy Loss: 0.0598, Value Loss: 0.0044
Epoch 4/5, Loss: 0.0605, Policy Loss: 0.0568, Value Loss: 0.0037
Epoch 5/5, Loss: 0.0594, Policy Loss: 0.0555, Value Loss: 0.0038
Completed iteration 9
Starting iteration 10/10
Generating self-play games...
Starting self-play game 1/10
  Move 10, State: rnbqk2r/ppppnp1p/3bp1p1/8/5P2/2P1P3/PP1PB1PP/RNBQK1NR w KQkq - 0 6
  Move 20, State: rnbqk2r/1p1pbp1p/4p1p1/p1p5/1P3P2/P1PPB3/4B1PP/RN1Q1KNR w kq - 1 11
  Move 30, State: r1b1k3/1p1pbprp/2n1p1p1/q1p5/1p3P1P/P1PP2PN/4B3/RNBQ1KR1 w q - 7 16
  Move 40, State: rqb1k3/n2pb1rp/1p2p1p1/2p2p2/Qp3P1P/P1PP2P1/R3B1R1/1NB2K1N w q - 0 21
  Move 50, State: r1bk4/n3br1p/1p1ppp1p1/5p2/1pp2B1P/PQPP2P1/R3BKR1/1N5N w - - 0 26
  Move 60, State: r1b5/n1kr3p/1p1bp3/5pp1/1pp4P/PQPP1KP1/2R1B1R1/1N4BN w - - 4 31
  Move 70, State: 2b5/rk4rp/1B1bp3/1n3pp1/1pp4P/P1PP1KP1/2QR1R1/1N5N w - - 5 36
  Move 80, State: r1b5/7p/1k1bp2r/1n3pP1/1PpP4/P1QR1KP1/4B1R1/1N5N w - - 3 41
  Move 90, State: 2b5/k1n4p/3bp2P/8/1P1p1p2/N2p1KP1/2Q1B1R1/7N w - - 0 46
  Move 100, State: k7/2n4p/b3p2P/8/1Q1P1p2/3p1KP1/4BR2/4N2N w - - 3 51
  Move 110, State: 1k6/1b1n2Qp/4p2P/3P4/5P2/3p1KN1/4BR2/4N3 w - - 1 56
  Move 120, State: k7/7p/1nb1p2P/7Q/5P2/3pKR2/4B3/4NN2 w - - 6 61
  Move 130, State: k7/7p/4p2P/5Q2/5P2/n3Kb2/3pB1N1/5N2 w - - 0 66
  Move 140, State: 8/k6p/7P/4p2Q/2n1KP2/5BN1/8/2r1N3 w - - 2 71
  Move 150, State: 304/k6p/7P/7B/2nK1p2/7r/6N1/8 w - - 2 76
  Game finished with outcome: 0
```

Figure 4. Training feedback of the 10th iteration

The figure above has been captured after iteration 10. Our total loss has dropped to 0.0594 and our policy loss is stable and low. Our value loss is small which indicates either the model has really high confidence when making predictions or it overfits. The game after iteration 10, unlike the games we have seen, does not include any random play. Both players are aware of material preservation, pawn advancement, and endgame ideas. It maintains pawn islands, leverages rooks in open files, and protects the king with all signs of a learning-based strategy. Looking at the FEN structure, in the opening both players prioritize the development of their pieces: knights and bishops are put in the center of the board and we can see early pawn movements on both sides. In mid-game we see trading of the major pieces which is a sign that players have the knowledge of mid-game and pursuit to the endgame. The endgame model makes reasonable pawn pushes, maintains kingside defenses, no aimless wandering.

### 4.2 Evaluation of Xiangqi Alpha-Zero Model

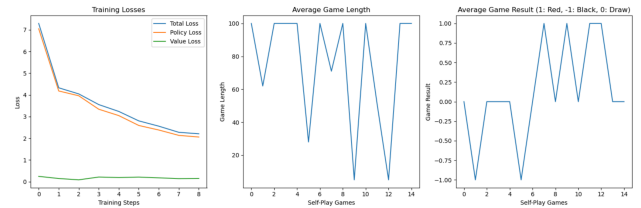

Figure 5. Evaluation result of Xiangqi

Over three rounds of training, we can already observe clear signs of learning: the average length of a game decreased from 92.4 to 60.8 and then rose slightly to 71.2, which means that the model is beginning to make more memorized moves while still searching. Also, the average reward transformed from negative in the first iteration (–0.20) to positive in the third iteration (+0.40), which indicates that the self-play outcome of the model is improving since it starts to favor more efficient strategies. On the training side, value and policy losses always decreased step by step, policy loss being the dominant constituent—this is in accord with typical AlphaZero behavior where policy network drives learning and the value head acts to stabilize. But in testing, the model would always draw every match (0 wins, 0 losses, 5 draws a round). Overall, the model shows stable convergence and improving self-play behavior.

### 4.2.1 Evaluation of Supervised Pre-training Model

In this section, we present excerpts from the opening moves of the first self-play game conducted with the pretrained model and, for comparison, one played with a randomly initialized model. These examples demonstrate how the pretrained policy influences early move selection.

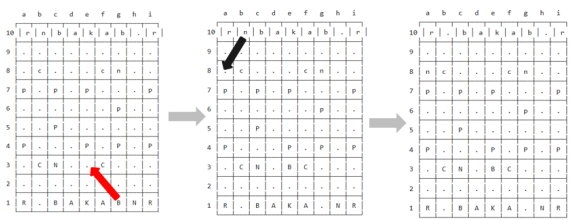

Figure 6. Movement without pretrained policy

The figure above illustrates the model's move choices when no pretrained model is used. Red arrows indicate Red's moves, and black arrows indicate Black's. As you can see, both sides are merely making random legal moves without any discernible offensive or defensive intent.

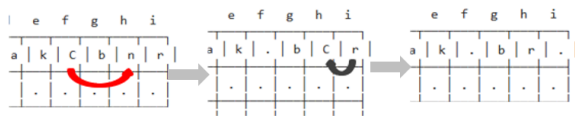

Figure 7. Movement with pretrained policy

By contrast, the figure above shows a simple attack–defense sequence that emerged after applying the pretrained model. Red's cannon (denoted "C") delivers a check by first capturing Black's knight to the right, forcing the Black general into check. To resolve this, Black captures Red's cannon with a chariot, thereby saving their general. This demonstrates that the model has learned a basic ability to judge straightforward attack and defense sequences. However, it is worth noting that, in human play, such a piece‑for‑piece trade ("duizi") is generally unfavorable for Red—indicating that the model still lacks a global, high‑level positional evaluation capability. This aspect will need to be further honed through continued self‑play training.

### 4.2.2 Evaluation of Final Model

After pre-training and self-playing part we load our model to play one last game to observe the success of the model. The final game consisted of the following moves ***f3-f10 - f8-f5 - b3-b4 -b10-c8 - f10-h10 - i10-h10 - b4-b2 - h10-h2 -i1-i3 - h2-e2.*** From the very early moves, the model shows a sense of rapid piece development, with a focus on developing chariots (Rooks) and cannon which are both crucial in the Xiangqi game. Red begins with f3-f10, a cannon hop to pressurize Black's mid-rank, and Black immediately reacts to this aggression with a cannon to f5. These are common aggressive opening strategies in real games, suggesting the policy has picked up helpful early-game patterns through self-play. Red rapidly deployed both chariots: f10-h10 and i1-i3, and Black mirrors and then swung h10-h2 to h2-e2, effectively swapping board space for rapid placement. But the one warning sign that catches the eye is Black's chariot retreat to the second rank (h10-h2) followed by diagonal move h2-e2), which gives Red a lot of initiative and does not directly challenge Red's structure. While Red's play seems more proactive, the b4-b2 (advancing the cannon to the back even more) maneuver is passive and may be symptomatic of state evaluation confusion or overfitting to draw-oriented positions. Similarly, Black's knight advance to c8 is standard but not accompanied by much centralization or support. Additionally, neither of them has advanced their pawns forward significantly, nor advanced the knights to maximum capacity which is a sign that the model lacks multi-phase planning.

## 5. Conclusion

In spite of playing under limited computational resources and time constraints, we were successful in implementing an AlphaZero-like reinforcement learning system for Xiangqi and for international chess. We first applied supervised pretraining on human game records to bootstrap a strong initial policy–value network, and then integrated it with Monte Carlo Tree Search for self-play. Our system learned to play legal, strategic games through this two-stage approach and demonstrated consistent improvement in move selection, positional awareness, and game outcomes across rounds. While it is not yet achieving expert-level play, the model avoids major strategic errors, mobilizes key pieces early, and exhibits clear learning patterns for foundational principles. Given the unique complexity of Xiangqi and the challenge of building a customized environment from scratch, our results reaffirm the effectiveness of this methodology.

## 6. Contribution of Team Members

| TEAM MEMBERS | CONTRIBUTION | MAIN WORK |
| --- | --- | --- |
| **Berk Yilmaz** | 1/3 | Alphazero chess and Xiangqi implementation, report writing |
| **Jinsong Liu** | 1/3 | creating the dataset and improve MCTS, report writing |
| **Junyu Hu** | 1/3 | supervised pre-training model implement, Report Writing |